\documentclass[10pt,twocolumn,letterpaper]{article}

\usepackage{lineno}
\usepackage{authblk}
\usepackage[pagenumbers]{cvpr} 

\definecolor{cvprblue}{rgb}{0.21,0.49,0.74}
\usepackage[pagebackref,breaklinks,colorlinks,allcolors=cvprblue]{hyperref}
\usepackage{algorithm}
\usepackage{algpseudocode}
\usepackage{amsmath}
\usepackage{xcolor}
\usepackage{cuted}
\usepackage{capt-of}
\usepackage{colortbl}
\usepackage{bm}

\definecolor{customgreen}{RGB}{34, 139, 34}

\def\paperID{5178} 
\def\confName{CVPR}
\def\confYear{2025}

\title{Single Trajectory Distillation for Accelerating Image and Video Style Transfer}

\author[1]{Sijie Xu}
\author[1,2]{Runqi Wang}
\author[1]{Wei Zhu}
\author[1]{Dejia Song}
\author[1]{Nemo Chen}
\author[1]{Xu Tang}
\author[1]{Yao Hu}
\affil[1]{Xiaohongshu}
\affil[2]{ShanghaiTech University}

\begin{document}
\maketitle
\begin{strip}\centering
    \includegraphics[width=\textwidth]{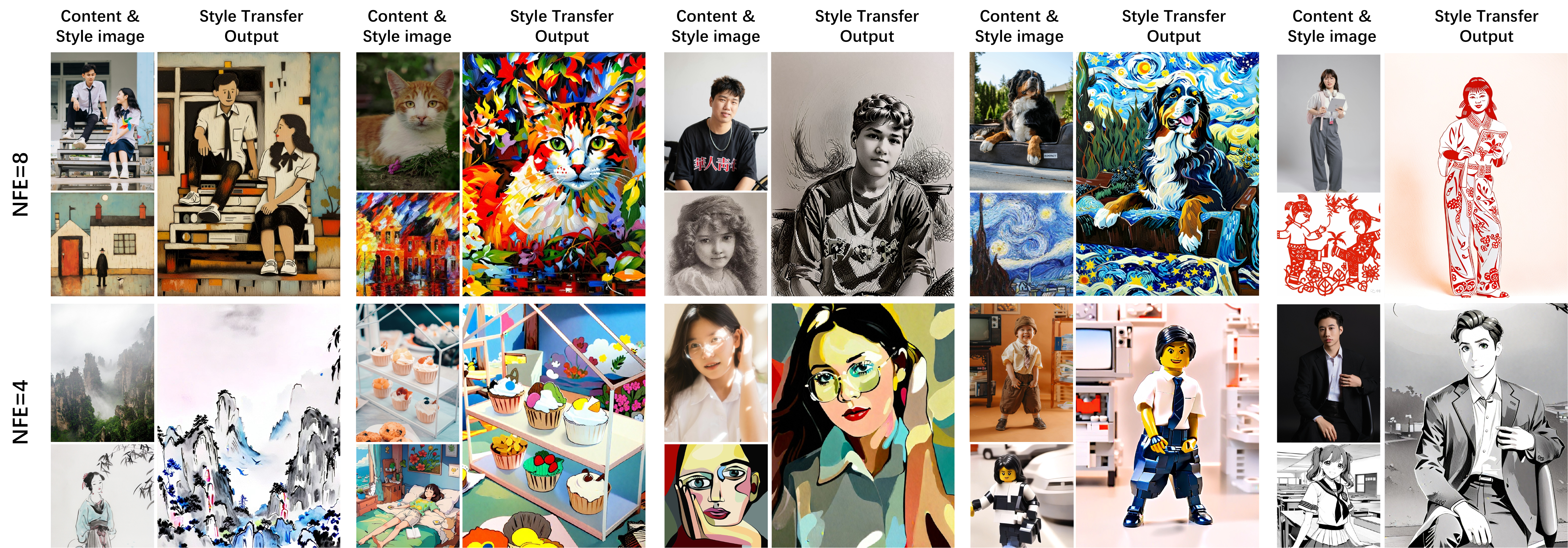}
    \captionof{figure}{Visualization of Results. Stylization examples of our method at the number of function evaluations (NFEs) 8 and 4.}
    \label{fig:fig1}
\end{strip}

\begin{abstract}
Diffusion-based stylization methods typically denoise from a specific partial noise state for image-to-image and video-to-video tasks.
This multi-step diffusion process is computationally expensive and hinders real-world application.
A promising solution to speed up the process is to obtain few-step consistency models through trajectory distillation.
However, current consistency models only force the initial-step alignment between the probability flow ODE (PF-ODE) trajectories of the student and the imperfect teacher models. 
This training strategy can not ensure the consistency of whole trajectories.
To address this issue, we propose single trajectory distillation (STD) starting from a specific partial noise state. We introduce a trajectory bank to store the teacher model's trajectory states, mitigating the time cost during training. Besides, we use an asymmetric adversarial loss to enhance the style and quality of the generated images. 
Extensive experiments on image and video stylization demonstrate that our method surpasses existing acceleration models in terms of style similarity and aesthetic evaluations.
Our code and results will be available on the project page: \url{https://single-trajectory-distillation.github.io/}.

\end{abstract}    
\section{Introduction}
Diffusion models have become crucial in generating and editing images and videos. Specifically, the method of perturbing and denoising, as described in SDEdit~\cite{meng2021sdedit}, is widely utilized for image and video editing tasks, particularly for stylization.
SDEdit-based methods first perturb the original image with a suitable amount of noise and then iteratively denoise the partial noise state.
Nonetheless, the slow sampling speed associated with diffusion models comes with challenges, particularly in video editing applications such as video stylization.
Acceleration methods can generally be divided into two categories: consistency model-based methods~\cite{luo2023latent,kim2023consistency,ren2024hyper} and non-consistency model-based methods~\cite{sauer2025adversarial,podell2023sdxl}. Consistency models~\cite{song2023consistency} seek to develop a distillation method exhibiting self-consistency.
The training objective for these models can be summarized as follows: $ \bm{f}(\bm{x}_t, \bm{x}_s) = \bm{f}(\bm{x}_{t'}, \bm{x}_s)$, where \( s < t' < t \in [0, T] \). Here, \( \bm{x}_{t'} := \bm{f}_\phi(\bm{x}_t, \bm{x}_{t'}) \), with \( \bm{f}(\cdot) \) representing the student network that is being trained and \( \bm{f}_\phi(\cdot) \) indicating the teacher network. This constraint ensures that the student model remains self-consistency along the trajectory defined by the teacher model.

Existing distillation methods based on consistency models apply the forward diffusion process to obtain $\bm{x}_t$, as depicted in Figure \ref{fig:fig2} (a).  This training strategy may cause misalignment along the entire trajectory, as it relies solely on initial-step constraints with an imperfect teacher model.
In detail, a random value for $t\in[0, T]$ is chosen, and $\bm{x}_t$ is obtained through the forward process, placing $\bm{x}_t$ on the forward diffusion trajectory during training.
The reverse trajectory with imperfect teacher diverges from the forward trajectory and the student model fits only the initial step of the teacher model's trajectory starting from time $t$, which makes the distillation suboptimal, amplifying the error across the entire trajectory.

We observe that a common approach in image and video stylization tasks is to add noise at a specific level and then denoise from this partial noise state. The sample distribution consistently evolves along the trajectory set generated by the teacher model, starting from the point on the forward trajectory at $\tau_\eta$.
In Figure \ref{fig:fig2} (b), we propose a single-trajectory distillation method that begins with $\tau_\eta$. For any $\bm{x}_t$ where $t\in[0, \tau_\eta]$, we obtain each initial state $\bm{x}_t$ by denoising $\bm{x}_{\tau_\eta}$ using the teacher model. 
Our goal is to achieve self-consistency throughout the teacher model's complete denoising trajectory, which aligns with the inference process.

Direct multi-step inference during training can significantly slow down the process. Therefore, this paper also proposes using a trajectory bank to store trajectory intermediate states, which maintains the input randomness without increasing the computational cost during training.
We present asymmetric adversarial loss that applies constraints between generated samples and real images with different noise levels to improve image saturation and decrease texture noise.
We constructed a test dataset of images and videos to assess the effectiveness of single-trajectory distillation in image and video stylization. Our method performs superiorly on stylization tasks compared to existing state-of-the-art techniques. The main contributions of this paper are as follows:

\begin{figure}[t]
  \centering
   \includegraphics[width=1.0\linewidth]{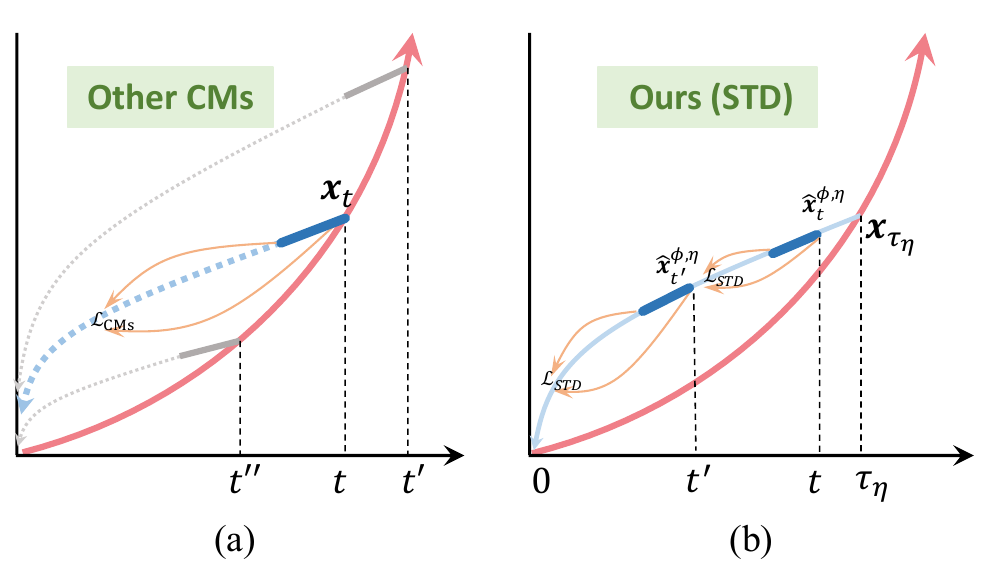}

   \caption{Comparison with other distillation schemes. (a) represents other distillation schemes where  $\bm{x}_t$  is obtained by adding noise to  $\bm{x}_0$, fitting only the initial portions of multiple PF-ODE trajectories. (b) represents our single-trajectory distillation scheme, where  $\bm{x}_t$  is derived by denoising from  $\bm{x}_{\tau_\eta}$, fitting a complete single trajectory starting from  $\bm{x}_{\tau_\eta}$.}
   \label{fig:fig2}
\end{figure}

\begin{enumerate}
    \item We propose a single-trajectory distillation method that provides consistency distillation with reduced error of the whole trajectory for partial noise editing tasks.
    \item We introduce a trajectory bank, which stores intermediate states of trajectories, allowing for more efficient training in single-trajectory distillation.
    \item We implement an asymmetric adversarial loss using DINO-v2 to enhance the style and quality of generated images.
\end{enumerate}

\begin{figure*}[t]
  \centering
   \includegraphics[width=1.0\linewidth]{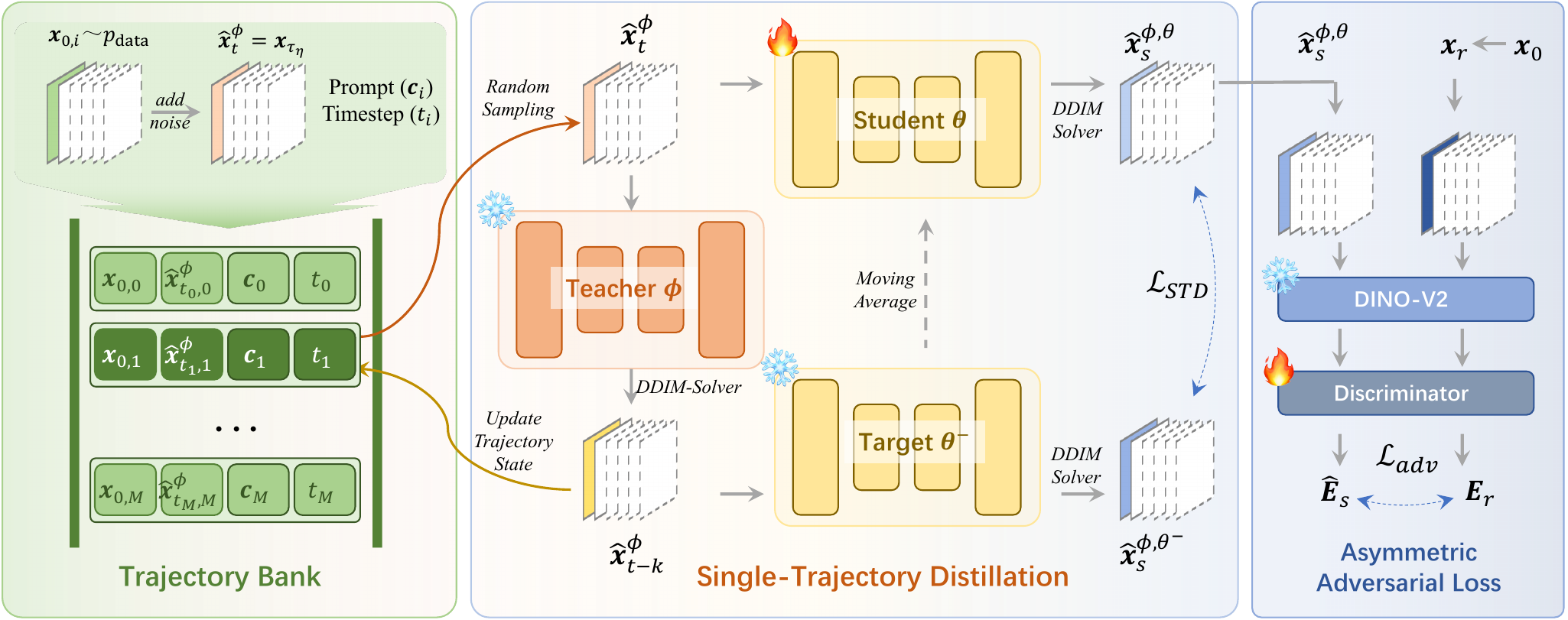}

   \caption{The diagram illustrates the single-trajectory distillation algorithm based on stable diffusion. On the left side is the trajectory bank, which manages samples from the teacher model's trajectory. Random samples $\bm{x}_t$ are drawn from this bank for training, and the sample states are updated after each one-step sampling by the teacher model to avoid repeated sampling and minimize time consumption. In the center, we present the single-trajectory distillation for image and video consistency distillation training. Here, only the student model is trained to align with the teacher model's trajectory. On the right side is the asymmetric adversarial loss component. The adversarial loss is based on DINO-v2, comparing the student model's prediction at timestep $s$ with the noisy ground truth at timestep $r$, where $r < s$. This approach improves the style and image quality.}
   \label{fig:fig3}
\end{figure*}

\section{Preliminaries}
\subsection{Diffusion Model}
\paragraph{Diffusion Model} Diffusion models~\cite{ho2020denoising} simulate the gradual addition of noise to data over several steps. During the generation phase, this process is reversed by incrementally removing the noise, ultimately resulting in realistic samples. From the viewpoint of score-based models~\cite{song2020score}, the forward process can be represented as a stochastic differential equation (SDE) that describes the transformation from the data distribution to a standard Gaussian noise distribution. This can be expressed as:
\begin{equation}
  d\bm{x}_t =\mu(t)\bm{x}_tdt+v(t)d\bm{w}_t
  \label{eq:1}
\end{equation}
For the reverse process, Song et al.~\cite{song2020score} demonstrated that there exists a probability flow ordinary differential equation (PF-ODE) that has the same boundary probability density as the forward SDE, expressed as:
\begin{equation}
  \frac{d\bm{x}_t}{dt}=\mu(t)\bm{x}_t-\frac{1}{2}v(t)^2\nabla_{\bm{x}}\log p_t(\bm{x}_t)
  \label{eq:2}
\end{equation}
In this context, $\nabla_{\bm{x}}\log p_t(\bm{x}_t)$ denotes the score the denoising model can predict. Various numerical methods for solving the PF-ODE can be developed by designing different solvers, such as DDIM-Solver \cite{song2020denoising}, DPM \cite{lu2022dpm}, DPM++ \cite{lu2022dpm++}, and others.

\paragraph{Trajectory} A trajectory visually represents the transformation of samples or sample distributions during noise addition and removal processes, a concept that has rarely been explicitly defined in prior research~\cite{song2020score,kim2023consistency,zheng2024trajectory}. A point on the trajectory is understood as the state of a specific sample after adding or removing a certain amount of noise, which is unsuitable for analyzing the distribution of trajectory samples. To improve understanding in the following discussions, we define a point on the trajectory as the boundary probability density at a specific timestep. This boundary probability density can be estimated using the sample distribution.
The forward diffusion SDE trajectory can be expressed as:
\begin{equation}
  \bm{x}_0\rightarrow \bm{x}_T := \{ p(\bm{x}_0, \bm{\epsilon}, t) \mid t\in [0, T] \}
  \label{eq:3}
\end{equation}
The reverse diffusion trajectory is expressed as:
\begin{equation}
  \bm{x}_T\rightarrow \hat {\bm{x}}_0^\phi := \{ q(\bm{\Phi}(\hat {\bm{x}}_t^\phi, T, t;\phi), \bm{z}, t)\mid t\in [0, T]\}
  \label{eq:4}
\end{equation}
where $\bm{\epsilon}, \bm{z} \sim \mathcal{N}(0, I)$, $\bm{\Phi}(\cdots;\phi)$ denotes an ODE solver with a denoising model $\phi$. When the model is well pre-trained and $\bm{x}_T \sim \mathcal{N}(0, \bm{I})$, then $\bm{x}_0 \equiv \hat {\bm{x}}_0^\phi$.

\paragraph{Partial Noise Editing} Due to the powerful generative capabilities and diversity of diffusion models, adding partial noise followed by denoising has become a common image and video editing method~\cite{meng2021sdedit,brooks2023instructpix2pix, duan2024diffutoon}. This method maintains the primary structure of the original image by retaining some of its information, with the denoising strength $\eta$ commonly used to regulate the extent of editing. The corresponding trajectory can be represented as:
\begin{equation}
  \bm{x}_{\tau_{\eta}}\rightarrow \hat {\bm{x}}_0^\phi := \{q(\bm{\Phi}(\hat {\bm{x}}_t^\phi, \tau_\eta, t;\phi), \bm{x}_{\tau_\eta}, t)\mid t\in [0, \tau_\eta]\}
  \label{eq:5}
\end{equation}
\begin{equation}
  \bm{x}_{\tau_\eta} = \mu_{\tau_\eta} \bm{x}_0 + v_{\tau_\eta} \bm{z}, \bm{z}\sim \mathcal{N}(0, \bm{I})
  \label{eq:6}
\end{equation}
where $\tau_\eta = \eta \cdot T$, and $T$ represents the total timesteps of the diffusion model. During the forward diffusion process, the mean and standard deviation at the timestep $\tau_\eta$ are denoted as $\mu_{\tau_\eta}$ and $v_{\tau_\eta}$.

\subsection{Consistency Model}
Song et al. \cite{song2023consistency} proposed and demonstrated that the number of inference steps can be reduced to accelerate the process while maintaining the self-consistency of the diffusion model.
\begin{equation}
  \bm{f}(\bm{x}_t, t)=\bm{f}(\bm{x}_{t'}, t')\    \forall t, t' \in [0, T]
  \label{eq:7}
\end{equation}
We can use distillation methods to efficiently obtain a consistency model, specifically,
\begin{equation}
  \mathcal{L}_{CMs}:=||\bm{f}_\theta(\bm{x}_{t_{n+1}}, t_{n+1}, s) - \bm{f}_{\theta^-}(\hat{\bm{x}}_{t_n}^\phi, t_n, s)||_{2}^2
  \label{eq:8}
\end{equation}
where \( 0 = t_1 < t_2 < \cdots < t_N = T \), and \(n\) is uniformly distributed over the set \(\{1, 2, \cdots, N-1\}\). The expectation is calculated with respect to the distribution \(\bm{x}_0 \sim p_{\text{data}}\). The sample \(\bm{x}_{t_{n+1}}\) is generated according to a SDE, while \(\hat{\bm{x}}^{\phi}_{t_n}\) is determined using the equation \(\hat{\bm{x}}^{\phi}_{t_n} := \bm{\Phi}(\bm{x}_{t_{n+1}}, t_{n+1}, t_n; \phi)\). Here, \(\bm{\Phi}(\cdots; \phi)\) represents the update function associated with an ODE solver. To improve stability during the training process, the parameter \(\theta^{-}\) is introduced and updated using an exponential moving average (EMA) strategy. 

\section{Related Works}

\subsection{Style Transfer}
Image and video stylization involves adjusting the style while maintaining the original structure, ensuring consistency with the input. This process requires both content and style information. There are several methods to introduce styles, such as through training style modules like LoRA~\cite{hu2021lora}, text embeddings~\cite{gal2022image,li2024styletokenizer}, and image embeddings~\cite{ye2023ip}. Content can be incorporated either as a generation condition (e.g., CSGO~\cite{xing2024csgo}) or as a loss constraint, as in StyleDiffusion~\cite{wang2023stylediffusion}. A common approach involves adding partial noise to the original image, naturally preserving its content. This method, first proposed by SDEdit~\cite{meng2021sdedit}, has been widely adopted for style transfer due to its simplicity, efficiency, and effectiveness.

\subsection{Step Distilling from Trajectory}
Diffusion models have significantly impacted various fields, including image synthesis and editing. However, their primary limitation is the slow multi-step inference process. To address this, Song et al.~\cite{song2023consistency} proposed using consistency models to distill the PF-ODE trajectory, enabling one-step or few-step generation. Subsequently, methods like CTM~\cite{kim2023consistency} and TCD~\cite{zheng2024trajectory} reduced the upper error bound through arbitrary-to-arbitrary distillation, improving consistency models' performance. Techniques such as Hyper-SD~\cite{ren2024hyper} and PCM~\cite{wang2024phased} further minimized errors by executing small-step distillation across different stages while aligning the changes in training and inference steps. TDD~\cite{wang2024target} enhanced the distillation process by introducing randomness in the target step position. AnimateLCM~\cite{wang2024animatelcm}, MCM~\cite{zhai2024motion} have expanded the use of consistency models in video generation.
However, these approaches assume that the teacher model is perfectly trained and obtains the $\bm{x}_t$ samples by adding noise to $\bm{x}_0$. Imagine Flash~\cite{kohler2024imagine}, DMD2~\cite{yin2024improved}, and SDXL-Lightning~\cite{lin2024sdxl} acquire $\hat {\bm{x}}_t$ samples through reverse diffusion, but they do not address the time-consuming issue caused by multi-step inference during training.

In samples with partial noise added, each value of $\eta$ corresponds to a unique trajectory (refer to the Appendix for more details). We aim to extract only the trajectory related to a specific $\eta$ to minimize errors caused by an imperfect teacher model. The $\hat {\bm{x}}_t$ sample is generated through reverse diffusion from $\bm{x}_{\tau_\eta}$. To optimize the training process, we propose using a trajectory bank to mitigate the time-consuming aspects of training.
\section{Methods}
This section will explain the single-trajectory distillation method and the accelerated training strategy that utilizes the trajectory bank.

\subsection{Single Trajectory Distillation}

For a given noise strength $\eta$, Equation \ref{eq:5} can express the reverse process trajectory. For any arbitrary value of \(\eta\), all trajectories can be represented as the set $\{\bm{x}_{\tau_{\eta(i)}} \rightarrow \bm{x}_0^{\eta(i)} | \eta(i) \in [0,1]\}$. When the denoising model is perfectly trained, all trajectories within this set converge into a single path that is identical to the forward diffusion trajectory but in reverse: $\bm{x}_T \rightarrow \bm{x}_0$. The samples $\bm{x}_0$ obtained by denoising along this perfect trajectory will perfectly match the \(p_{\text{data}}\) distribution. However, it is important to note that no model can perfectly fit the training data distribution.

\noindent \textbf{Theorem}: \textit{In an imperfect denoising model, the PF-ODE trajectories originating from any two points on the forward diffusion path are inconsistent. This means that for any values $\eta, \eta'\in [0,1]$, the trajectories $\bm{x}_{\tau_\eta} \rightarrow \bm{x}_0^{\eta}$ and $\bm{x}_{\tau_{\eta'}} \rightarrow \bm{x}_0^{\eta'}$ are not equivalent.}

\noindent \textbf{Proof.}: For analytical convenience, we assume that the imperfect denoising model predicts noise $\bm{\epsilon}_\phi$, where $\bm{\epsilon} - \bm{\epsilon}_\phi < \bm{\delta}_\phi$, where $\bm{\epsilon}\sim\mathcal{N}(0, \bm{I})$. Suppose the denoising process employs a DDIM-Solver, denoted as $\bm{\Phi}(\cdots;\phi)$, and let $\tau_\eta = t, \tau_{\eta'} = s, s<t$. It suffices to prove that the forward sample $\bm{x}_s$ is not equal to the denoising sample $\hat {\bm{x}}_s^\phi$. From the forward diffusion process and the DDIM formulation, we can derive the following conclusions:
\begin{equation}
  \bm{x}_s - \hat {\bm{x}}_s^\phi < C_{t,s}\cdot \bm{\delta}_\phi
  \label{eq:4.1}
\end{equation}
\begin{equation}
  C_{t, s} = \frac{1}{\sqrt{\alpha_t}}(\sqrt{\alpha_s}\sqrt{1-\alpha_t} - \sqrt{\alpha_t}\sqrt{1-\alpha_s})
  \label{eq:4.2}
\end{equation}
The detailed derivation can be found in Supplementary Material \ref{proof}.

In tasks involving image and video stylization, inference typically starts with denoising from a predetermined noise level $\eta$. Distillation can be specifically tailored to the complete trajectory for a particular $\eta$, called single-trajectory distillation. This approach reduces error and improves alignment between the training and inference processes.
Distillation can be specifically tailored to the complete trajectory for a particular $\eta$, called single-trajectory distillation. This approach reduces error and improves alignment between the training and inference processes.
\begin{equation}
  \mathcal{L}_{STD}:=||\bm{f}_{\theta}(\hat{\bm{x}}_{t_{n+1}}^{\phi,\eta}, t_{n+1}, s) - \bm{f}_{\theta^-}(\hat{\bm{x}}_{t_n}^{\phi,\eta}, t_n, s)||_{2}^2
  \label{eq:4.3}
\end{equation}
\begin{equation}
  \hat{\bm{x}}_{t_{n+1}}^{\phi, \eta} = \bm{\Phi}(\bm{x}_{\tau_\eta}, \tau_\eta, t;\phi),\tau_\eta=\eta\cdot T
  \label{eq:4.4}
\end{equation}
\begin{equation}
  \hat{\bm{x}}_{t_n}^{\phi, \eta} = \bm{\Phi}(\hat{\bm{x}}_{t_{n+1}}^{\phi, \eta}, t_{n+1}, t_n;\phi)
  \label{eq:4.5}
\end{equation}
\begin{equation}
  \bm{x}_{\tau_\eta} = \alpha_{\tau_\eta}\bm{x}_0+\sigma_{\tau_\eta}\bm{\epsilon}, \bm{\epsilon} \sim \mathcal{N}(0, \bm{I})
  \label{eq:4.6}
\end{equation}

Based on previous research, a short distance between \( t \) and \( s \) can lower the upper bound of the distillation error while preserving the randomness of \( s \), thus improving the performance of the model. To achieve this, we sample \( s \) from an interval ranging from \( t \) to \( 0 \), as follows:
\begin{equation}
  s\sim \mathcal{U}[(1-\gamma)t, t], t\in[0, \tau_\eta]
  \label{eq:4.7}
\end{equation}

\subsection{Trajectory Bank}

For the training of STD, equation \ref{eq:4.4} requires multi-step ODE-Solver reverse diffusion, leading to increased training time. To address this issue, we propose the Trajectory Bank as equation \ref{eq:4.8}, which stores intermediate states of the teacher model along the reverse diffusion trajectory $\bm{x}_{\tau_\eta} \rightarrow \bm{x}_0^\eta$. By randomly sampling $\hat{\bm{x}}_t^\phi$ from the bank according to the sampling probability $\rho$ at each step, we can directly obtain the trajectory’s previous state at time $t$.  The trajectory bank can be defined as:

\begin{equation}
  \mathcal{B}:=\{(\bm{x}_{0,i}), \hat {\bm{x}}_{t,i}^\phi, \bm{c}_i, t_i)\mid i\in[0,M]\}
  \label{eq:4.8}
\end{equation}

As shown on the left of Figure \ref{fig:fig3}, the trajectory bank samples $\bm{x}_0$ from the dataset, which the forward SDE then perturbs to the noise level $\tau_\eta$. Both the sample and its corresponding prompt $\bm{c}_i$ are stored in the bank. During training, a random sample is drawn from the bank. After passing through the teacher model, it produces $\hat {\bm{x}}_{t_n}^{\phi, \eta}$ and the corresponding timestep to replace the sample in the bank, enabling sample updating. On the subsequent retrieval of this sample, the model directly processes it from $\bm{x}_{t_n}^{\phi, \eta}$ to $\bm{x}_{t_{n-1}}^{\phi, \eta}$, and this process repeats. When $t_n = 0$, the sample is popped from the bank, and a new $\bm{x}_{\tau_\eta}^{\prime}$ is added.

\begin{algorithm}[htbp]
\caption{Single Trajectory Distillation}
\label{alg:std}
\textbf{Input:} dataset $\mathcal{D}$, learning probability $\delta$, the update function of ODE-Solver $\bm{\Phi}(\cdots;\phi)$, EMA rate $\mu$, noise schedule $\alpha_t$, $\sigma_t$, number of ODE steps $N$, random guidance scale $\omega$, trajectory bank $\mathcal{B}$, bank size $M$, sampling from trajectory bank rate $\rho$, target step sample range rate $\gamma$, total time steps $T$ and strength $\eta$.\\
\textbf{Parameter:} initial model parameter $\theta$, discriminator parameter $\psi$ \\
\begin{algorithmic}[1]
\Repeat
    \State \textcolor{customgreen}{\# Prepare and Sample Input}
    \State Sample $(\bm{x}_0, c) \sim \mathcal{D}$, $\omega \sim \mathcal{U}[\omega_{min}, \omega_{max}]$
    \If{probability $<$ $\rho$ and $\mathcal{B}$ is not empty}
        \State $\bm{x}_0, \bm{x}_{t_{n+1}}, \bm{c}, t_{n+1} \sim \mathcal{B}(i), i\sim \mathcal{U}[0, M]$
    \Else
        \State $t_{n+1} = \eta \cdot T$
        \State $\bm{x}_{t_{n+1}} = \alpha_{t_{n+1}}\bm{x}_0 + \sigma_{t_{n+1}}\bm{\epsilon}, \bm{\epsilon}\sim\mathcal{N}(0, \bm{I})$
        \State $i \leftarrow \text{free\_index}(\mathcal{B}) \ \text{if} \ |\mathcal{B}| < M \ \text{else} \ \varnothing$
    \EndIf
    \State Sample $s\sim \mathcal{U}[(1-\gamma)t_{n+1}, t_{n+1}]$
    \State \textcolor{customgreen}{\# Consistency Distillation}
    \State $\hat{\bm{x}}_{t_n}^{\phi, \eta} \leftarrow (1 + \omega)\bm{\Phi}(\bm{x}_{t_{n+1}}, t_{n+1}, t_n, \bm{c}; \phi)$
    \State \hspace{16pt} $- \omega \bm{\Phi}(\bm{x}_{t_{n+1}}, t_{n+1}, t_n, \varnothing; \phi)$
    \State $\mathcal{L}_{\text{STD}} := \left\| \bm{f}_{\theta}(\bm{x}_{t_{n+1}}, t_{n+1}, s) - \bm{f}_{\theta^-}\left(\hat{\bm{x}}^{\phi, \eta}_{t_n}, t_n, s \right) \right\|_2^2$
    \State $\mathcal{L}_{adv}^G, \mathcal{L}_{adv}^D \leftarrow$ Eq. \ref{eq:4.8},\ref{eq:4.9}
    \State $\mathcal{L}(\theta, \theta^{-}, \psi; \phi, D_\psi)\leftarrow \mathcal{L}_{\text{STD}}+\lambda_{adv}\mathcal{L}_{adv}^G$
    \State $\theta \leftarrow \theta - \delta \nabla_{\theta} \mathcal{L}(\theta, \theta^{-}, \psi; \phi, D_\psi)$
    \State $\theta^{-} \leftarrow \text{sg}\left(\mu \theta^{-} + (1-\mu)\theta\right)$
    \State $\psi\leftarrow \psi - \delta_{D_\psi}\nabla_\psi\mathcal{L}_{adv}^D$
    \State \textcolor{customgreen}{\# Update Trajectory Bank}
    \If{$i$ is not $\varnothing$}
        \State $\mathcal{B}(i) \leftarrow \hat{\bm{x}}_{t_n}^{\phi, \eta}, t_n$
    \EndIf
\Until{convergence}
\end{algorithmic}
\end{algorithm}
\begin{figure*}[htbp]
  \centering
   \includegraphics[width=1.0\linewidth]{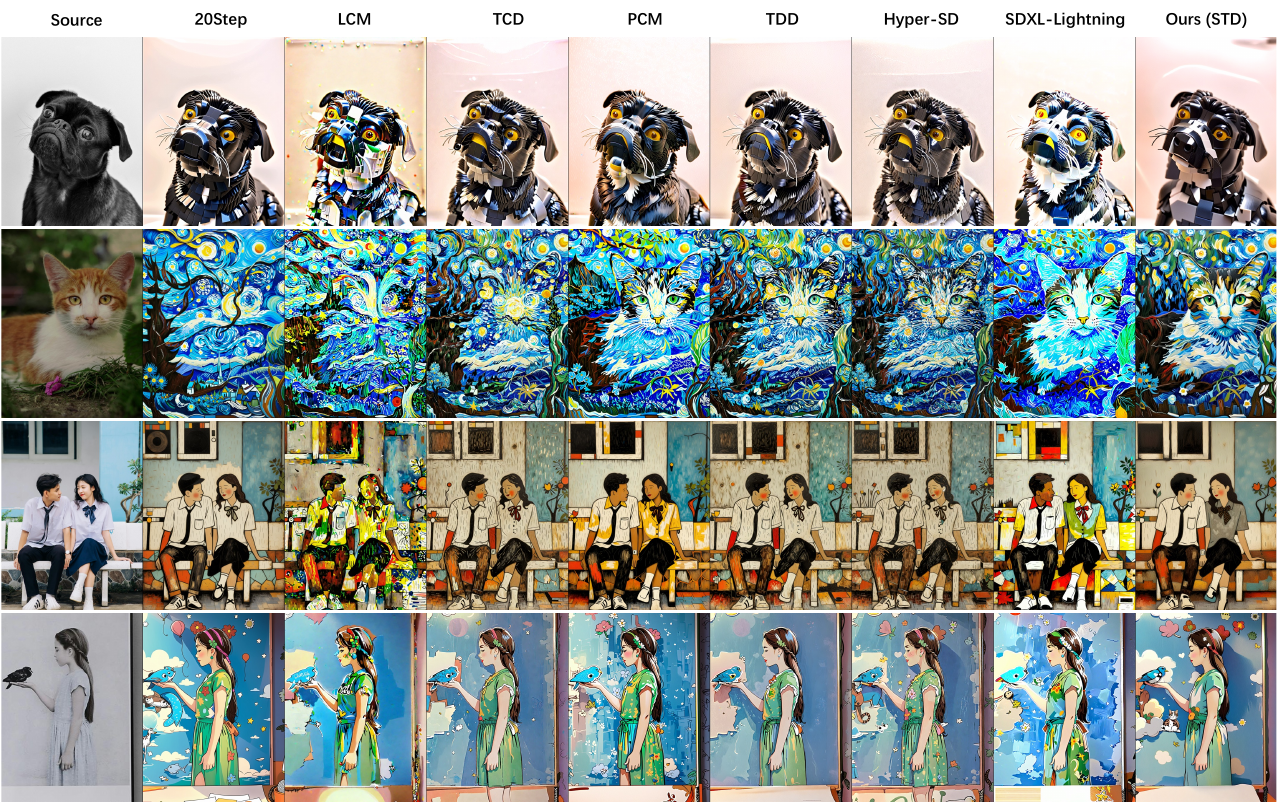}

   \caption{Comparison of Experimental Results. The figure shows some comparison examples among our method, STD, and other acceleration methods, including LCM, TCD, PCM, TDD, Hyper-SD, and SDXL-Lightning. On the left, we also present the original image and the result obtained using a 20-step Euler Solver. All methods are evaluated under the settings of CFG=6 and NFE=8.}
   \label{fig:fig4}
\end{figure*}

\subsection{Asymmetric Adversarial Loss}


We observe that images generated via consistency distillation often exhibit noticeable speckle noise. Many approaches have employed adversarial loss functions to enhance the stability and quality of the generated images~\cite{wang2024phased,wang2024target,sauer2025adversarial,zhai2024motion}. In particular, Wang et al.~\cite{wang2024phased} utilized the entire denoising model as the discriminator, whereas Sauer et al.~\cite{sauer2025adversarial} trained a smaller discriminator using DINO-v2~\cite{oquab2023dinov2} features. We adopt the latter approach for two main reasons: 
\begin{itemize}
    \item Our method aims to learn the trajectory $\bm{x}_{\tau_\eta} \to \hat {\bm{x}}_0^\phi$, reaching the distribution of $\hat {\bm{x}}_0^\phi$, rather than the original data distribution $\bm{x}_0$. Utilizing DINO-v2 features allows for semantic constraints, avoiding pixel-level constraints. 
    \item Employing a DINO-v2 discriminator when performing video distillation offers greater memory efficiency, enabling improved processing of video data without significantly increasing memory usage.

\end{itemize}
We propose an asymmetric adversarial loss function to introduce randomness and reduce speckle noise. In contrast to MCM~\cite{zhai2024motion}, which constrains $\hat {\bm{x}}_0^{\theta}$ to the real image $\bm{x}_0$, we utilize an asymmetric constraint between $\hat {\bm{x}}_s^{\phi,\theta}$ and $\bm{x}_r$, where $0 \leq r < s$.

Specifically, we denote $\hat{\bm{x}}_s^{\phi,\theta} = \bm{f}_\theta(\hat {\bm{x}}_t^{\phi,\eta}, t, s;\phi)$, which represents the sample obtained by the student network performing reverse diffusion from time step $t$ to $s$ using the ODE-Solver. For video samples, we follow the MCM approach by randomly sampling $l$ frames, denoted as $\{\hat{\bm{x}}_{s, 0}^{\phi,\theta},\dots, \hat{\bm{x}}_{s, l}^{\phi,\theta}\}$. For convenience, we use the same notation for image samples, where $l=1$. The adversarial loss can then be expressed as:
\begin{equation}
  \mathcal{L}_{\text{adv}}^G := -\mathbb{E}_{x_0, \bm{\epsilon}, t} \left[ \frac{1}{l} \sum_i D_\psi \left( \mathcal{F}(\hat{\bm{x}}_{s,i}^{\phi,\theta}) \right) \right]
  \label{eq:4.9}
\end{equation}
\begin{equation}
  \begin{aligned}
    \mathcal{L}_{\text{adv}}^D := &\ \mathbb{E}_{x_0, \bm{\epsilon}, t} \left[ \max \left( 0, 1 + \frac{1}{l} \sum_i D_\psi \left( \mathcal{F}(\hat{\bm{x}}_{s,i}^{\phi,\theta}) \right) \right) \right] \\
    &+ \mathbb{E}_{m_i} \left[ \max \left( 0, 1 - D_\psi \left( \mathcal{F}(\bm{x}_{r,i}) \right) \right) \right], r\in[0, s]
    \end{aligned}
  \label{eq:4.10}
\end{equation}
where $\mathcal{F}$ represents the DINO-v2 model, $D_\psi$ denotes the discriminator, $\psi$ represents the learnable parameters of the discriminator, and $\bm{x}_r$ refers to the sample obtained by adding noise to $\bm{x}_0$ for $r$ steps, with $r$ being randomly selected from the range $0$ to $t$. The final loss function can be expressed as:
\begin{equation}
  \mathcal{L}=\mathcal{L}_{STD}+\lambda_{adv}\mathcal{L}_{adv}^G
  \label{eq:4.11}
\end{equation}
where the $\lambda_{adv}$ represents the adversarial loss weight.

\section{Experiments}
\subsection{Implementations}
\paragraph{Datasets} 
We selected the video dataset Pandas70M~\cite{chen2024panda70m} as our training set to accommodate both the video and image distillation in our training process. When training the image diffusion distillation of STD, we used the first frame of each video sample as the training data, while for the video version, the entire video frame sequence was employed. We constructed some images and videos as test sets like previous works~\cite{chen2024artadapter, wang2023stylediffusion,duan2024diffutoon,ku2024anyv2v} for the test set, which included 100 images and 12 video clips of diverse subjects, containing pets and people in different poses and landscapes. We also selected 15 representative iconic reference style images for style injection, following the principle of covering a wide range of distinct styles, as adopted in existing works~\cite{chen2024artadapter, li2024styletokenizer, xing2024csgo}.
\paragraph{Metrics}
For image stylization evaluation, we primarily use the CSD score~\cite{somepalli2024measuring, xing2024csgo} to measure the style similarity with the reference style image, and the LAION-Aesthetics Predictor~\cite{burger2023laion} to assess the aesthetic quality of the stylized image. We sample video frames at regular intervals to calculate the average CSD and aesthetic scores. Simultaneously, we assess warping error~\cite{lei2020dvp,lei2022deep} to evaluate the temporal consistency among video frames after stylization.

\paragraph{Training}
For the training of the image-based STD model, we choose Stable Diffusion XL (SDXL)~\cite{podell2023sdxl} as the base model, and for the video-based version, we use the Animatediff-SDXL~\cite{guo2023animatediff} model, where the motion module is Hotshot-XL~\cite{mullan2023hsxl}. We employ a 50-step DDIM solver~\cite{song2020denoising} as the ODE solver $\bm{\Phi}$ and train a LoRA~\cite{luo2023lcm} with a rank of $64$ to replace fine-tuning the entire U-Net. The size of the STD trajectory bank is set to $4$, and $\eta$ is set to $0.75$.  We also compare different values of $\eta$ in Supplementary Material \ref{comp}. The learning rate for the diffusion model is set to $5 \times 10^{-6}$, the learning rate for the discriminator is set to $5 \times 10^{-5}$, and the batch size is $128$. These settings are consistent with MCM~\cite{zhai2024motion}. We use the Adam optimizer and train for $12k$ steps. Additional details regarding training, sampling, and testing will be available in the Supplementary Material \ref{sec:imple}.

\subsection{Style Transfer Acceleration}
We use IP-Adapter~\cite{ye2023ip} to inject styles and follow the approach of InstantStyle~\cite{wang2024instantstyle} by injecting image embeddings only into the style block of SDXL. We evaluate both image and video stylization and compare our method with existing open-sourced acceleration approaches with SDXL models, including LCM~\cite{luo2023latent}, TCD~\cite{zheng2024trajectory}, PCM~\cite{wang2024phased}, TDD~\cite{wang2024target}, Hyper-SD~\cite{ren2024hyper}, and SDXL-Lightning~\cite{lin2024sdxl}. Additionally, we include the MCM~\cite{zhai2024motion} model for the video version as a comparison. Due to the MCM merely supporting models trained on SD1.5, we trained MCM with SDXL model using the same training dataset as STD.

\paragraph{Image Style Transfer} We evaluated the performance of various acceleration models at a noise strength of 0.75. As shown in Table \ref{tab:comparation}, our method outperforms existing methods in terms of style similarity and aesthetic quality. Figure \ref{fig:fig4} illustrates several comparison examples at eight steps across different methods. It can be observed that our results exhibit higher contrast, fewer extraneous details, and reduced noise compared to other methods. In the style transfer domain, reduced detail often results in cleaner and more aesthetically pleasing images.

\begin{table*}[htbp]
  \centering
  \caption{Comparison of objective image and video stylization metrics at different sampling steps.}
    \begin{tabular}{c|cccccc|ccc}
    \toprule
          & \multicolumn{6}{c|}{Image Style Transfer}     & \multicolumn{3}{c}{Video Style Transfer} \\
    Steps & \multicolumn{2}{c}{NFE=8} & \multicolumn{2}{c}{NFE=6} & \multicolumn{2}{c|}{NFE=4} & \multicolumn{3}{c}{NFE=8} \\
    Metrics & CSD   & Aesthetic & CSD   & Aesthetic & CSD   & Aesthetic & CSD   & Aesthetic & Warping Error \\
    \midrule
    LCM~\cite{luo2023latent}   & 0.426 & 4.578 & 0.399 & 4.427 & 0.366 & 4.210  & 0.419 & 4.210  & 0.320 \\
    TCD~\cite{zheng2024trajectory}   & 0.520  & 5.092 & \underline{0.505} & 5.040  & \underline{0.452} & 4.705 & 0.480  & 4.938 & 0.173 \\
    PCM~\cite{wang2024phased}   & 0.473 & 5.012 & -     & -     & 0.341 & 4.623 & 0.349 & 4.226 & 0.261 \\
    TDD~\cite{wang2024target}   & 0.509 & 5.156 & 0.497 & \underline{5.131} & \underline{0.452} & \underline{4.827} & 0.478 & \underline{4.965} & 0.199 \\
    Hyper-SD~\cite{ren2024hyper} & \underline{0.522} & \underline{5.163} & -     & -     & 0.400   & 4.322 & \underline{0.490} & \textbf{4.970} & \textbf{0.153} \\
    SDXL-Lightning~\cite{lin2024sdxl} & 0.447 & 4.678 & -     & -     & 0.297 & 4.332 & 0.309 & 3.944 & 0.234 \\
    MCM~\cite{zhai2024motion}   & -     & -     & -     & -     & -     & -     & 0.442 & 4.381 & 0.257 \\
    \rowcolor[rgb]{ .953,  .953,  .957} Ours (STD) & \textbf{0.554} & \textbf{5.190} & \textbf{0.525} & \textbf{5.176} & \textbf{0.489} & \textbf{4.896} & \textbf{0.507} & 4.953 & \underline{0.166} \\
    \bottomrule
    \end{tabular}%
  \label{tab:comparation}%
\end{table*}%

\begin{figure}[t]
  \centering
   \includegraphics[width=1.0\linewidth]{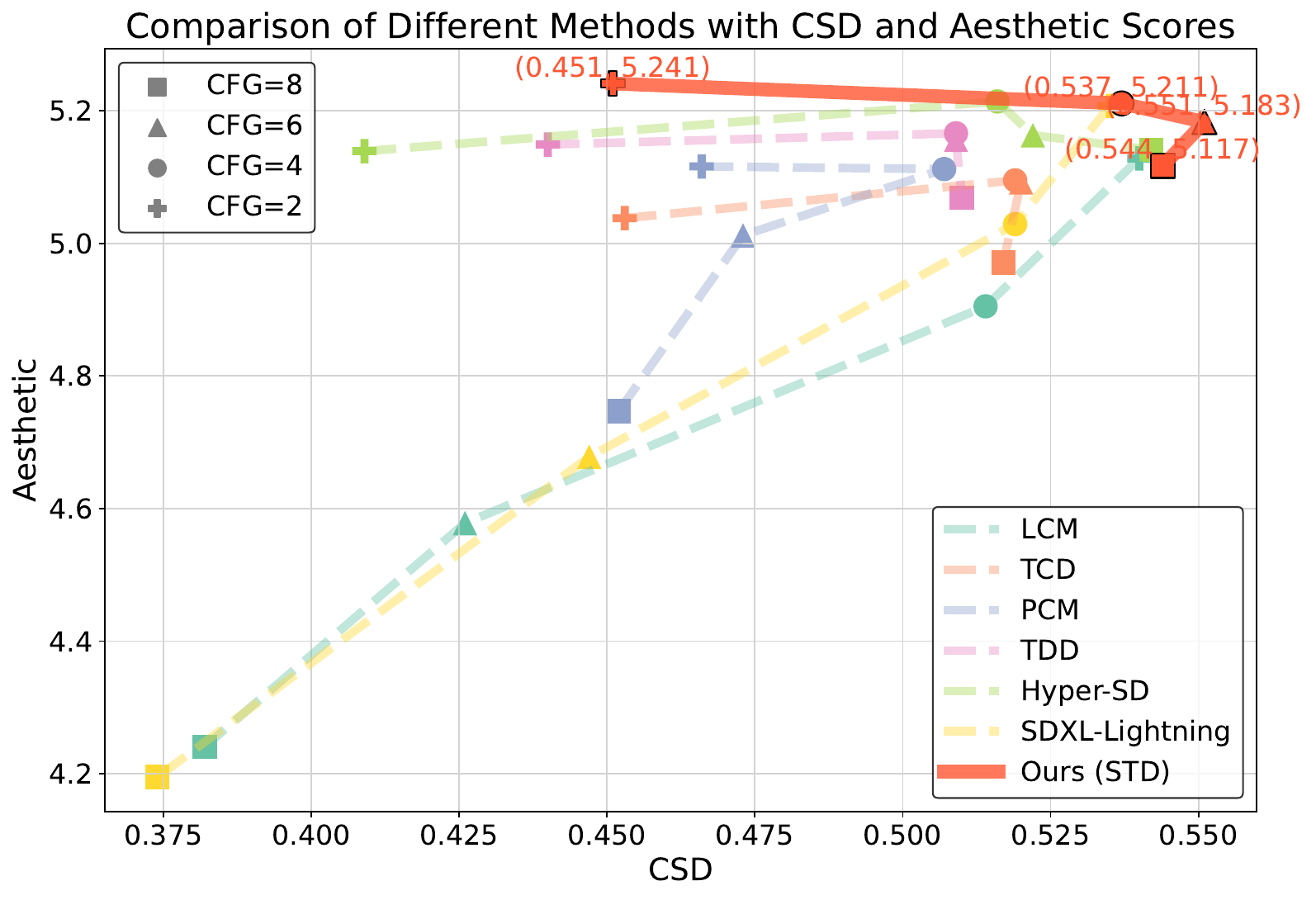}
   \caption{Line chart comparing methods under different CFG values. The horizontal axis represents the style similarity metric (CSD), and the vertical axis represents the aesthetic score. The chart shows the metric lines for our method and comparison methods at CFG values of 2, 4, 6, and 8, where closer proximity to the upper-right corner indicates better performance.}
   \label{fig:fig5}
\end{figure}

Previous work~\cite{wang2024phased,wang2024target} has demonstrated that classifier-free guidance (CFG) substantially impacts results. We evaluated each method with CFG values of 2, 4, 6, and 8 to facilitate a more fair comparison, as shown in Figure \ref{fig:fig5}. Our method exhibits robustness across different CFG values. We also present the results of the comparison method using the recommended CFG in the Supplementary Material \ref{comp}.

\paragraph{Video Style Transfer} In our video stylization comparison experiment, we limit the evaluation to eight steps of metrics due to the long processing time required for video testing. The results reveal that the inclusion of the motion module leads to a degradation in overall style and image quality. Our method still maintains better style preservation than others, with over-average aesthetic scores and inter-frame consistency. Additional examples of video are provided in Supplementary Material \ref{comp}.

\subsection{Ablation Study} 
Our ablation study primarily focuses on the two key proposed contributions: STD and asymmetric adversarial loss. Additionally, we discuss some of the associated hyperparameters.
As shown in Table \ref{tab:ablation_all}, we find that both STD and the asymmetric adversarial loss improve style similarity and aesthetic scores. The best results are achieved when both are used together.

\begin{table}[h]
  \centering
  \caption{Ablation study on single-trajectory distillation and asymmetric adversarial loss.}
    \begin{tabular}{cc|cc}
    \toprule
    STD   & Asymmetric Adv.  & CSD   & Aesthetic \\
    \midrule
          &       & 0.518 & 5.151 \\
    \checkmark   &       & 0.539 & 5.191 \\
          & \checkmark   & 0.536 & 5.164 \\
    \checkmark   & \checkmark   & \textbf{0.561} & \textbf{5.202} \\
    \bottomrule
    \end{tabular}%
  \label{tab:ablation_all}%
\end{table}%

\begin{figure}[h]
  \centering
   \includegraphics[width=1.0\linewidth]{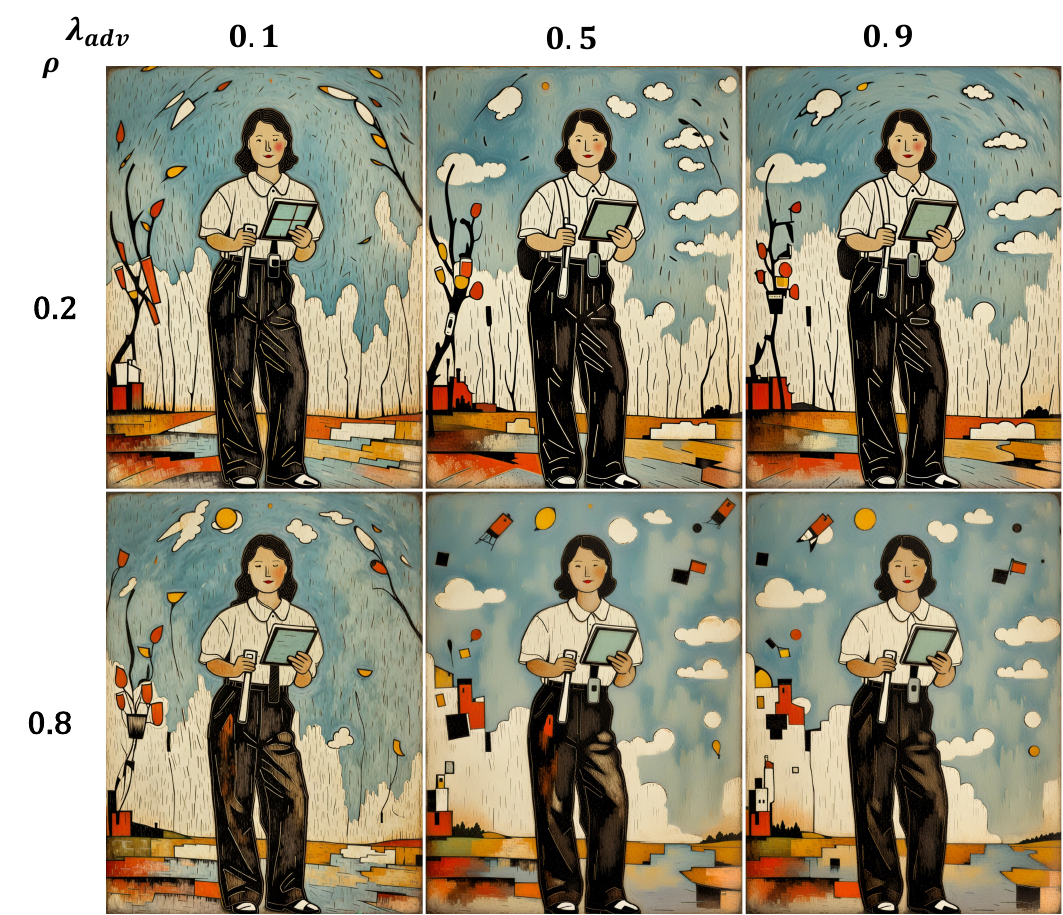}
   \caption{Ablation study on the strength of single-trajectory distillation and asymmetric adversarial loss. Each row represents different weights for the asymmetric adversarial loss, and each column represents varying sample probabilities for STD.}
   \label{fig:fig6}
\end{figure}

Additionally, we investigated the impact of the strengths in single-trajectory distillation and asymmetric adversarial loss. The corresponding hyperparameters are the trigger sampling probability  $\rho$  for single-trajectory distillation and the weight  $\lambda_{adv}$  for the asymmetric adversarial loss. As shown in Figure \ref{fig:fig6}, as  $\rho$  or  $\lambda_{adv}$  increases, the image's noise is significantly reduced, resulting in a cleaner overall appearance. Increasing the weight of the asymmetric adversarial loss also enhances image contrast and saturation, which reflects an intensification of style.

To further investigate the effectiveness of the asymmetric adversarial loss, we designed the experiment with different target time step constraints: the generated samples and real images at timestep $0$, $s$ and $r$, where $s$ is the target timestep of student model and $r$ is a random timestep between $0$ and $\tau_\eta$. All other training settings remain consistent, and we evaluate objective metrics on an 8-step sampling process, as shown in Table \ref{tab:ablation_gan}. Constraint at timestep $s$ instead of timestep $0$ results in higher style similarity and aesthetic scores. 
We attribute this to the fact that our single-trajectory distillation simulates the teacher model’s trajectory. There is still significant difference between the predicted $ \hat {\bm{x}}_0^{\phi,\theta}$ and the actual image $\bm{x}_0$ when denoising from a point on this trajectory to $ \hat {\bm{x}}_0^{\phi,\theta}$. 
Attempting to align these distributions too closely can misalign the learning target of the student model with the teacher’s trajectory, causing style and aesthetic degradation. 
When both the generated samples and real images are at step $s$, high-frequency details are effectively removed, reducing distribution alignment to semantic consistency and facilitating student learning. 

To explore the impact of the selection of random timestep $r$, we adopt asymmetric adversarial constraint between the gernerated samples $ \hat {\bm{x}}_s^{\phi,\theta}$ and real images $\bm{x}_r$. When the timestep $r$ of a real image $\bm{x}_r$ with noise satisfies $r < s$, the student model tends to remove more noise, resulting in a clearer image. Conversely, when the timestep $r$ satisfies $r > s$, the student model retains more noise, leading to speckle noise issues.

We also analyze the impact of different target timestep range rates $\gamma$ on the predicted images. Further details can be found in Supplementary Material \ref{ablation}.

\begin{table}[htbp]
  \centering
  \caption{Ablation study on input noise intensity for asymmetric adversarial loss.}
    \begin{tabular}{cc|cc}
    \toprule
    Pred. Timestep & GT. Timestep & CSD   & Aesthetic \\
    \midrule
    $0$     & $0$     & 0.514 & 5.026 \\
    $s$     & $s$     & 0.537 & \textbf{5.193} \\
    $s$     & $r$ $(r>s)$ & 0.422 & 4.773 \\
    $s$     & $r$ $(r<s)$ & \textbf{0.551} & 5.183 \\
    \bottomrule
    \end{tabular}%
  \label{tab:ablation_gan}%
\end{table}%

\section{Conclusion}
This paper proposes combining single-trajectory distillation with asymmetric adversarial loss to accelerate image and video stylization. 
We demonstrate that the PF-ODE trajectory varies with different noise intensities added along the forward SDE trajectory. 
Based on this, we introduce a single-trajectory distillation approach with a specific noise strength, and a trajectory bank, to effectively solve the misalignment between trajetories of training and inference. 
Furthermore, we employ an asymmetric adversarial loss to enhance image and video quality. 
Comparative experiments validate the superiority of our method in terms of style and aesthetics, and ablation studies confirm the effectiveness of each module. 
Our method extends beyond video stylization and could theoretically apply to all partially noised editing tasks, such as image inpainting, which is our future work.
{
    \small
    \bibliographystyle{ieeenat_fullname}
    \bibliography{main}
}

\clearpage
\setcounter{page}{1}
\maketitlesupplementary

\section{Theoretical Proof}
\label{proof}

\noindent \textbf{Theorem}: \textit{In an imperfect denoising model, the PF-ODE trajectories originating from any two points on the forward diffusion path are inconsistent. This means that for any values $\eta, \eta'\in [0,1]$, the trajectories $\bm{x}_{\tau_\eta} \rightarrow \bm{x}_0^{\eta}$ and $\bm{x}_{\tau_{\eta'}} \rightarrow \bm{x}_0^{\eta'}$ are not equivalent.}

\noindent \textbf{Proof.}: To prove that the denoising trajectories $\bm{x}_{\tau_\eta} \rightarrow \bm{x}_0^{\eta}$ and $\bm{x}_{\tau_{\eta'}} \rightarrow \bm{x}_0^{\eta'}$ starting from two different strengths on the SDE trajectory are different, it can be reduced to showing that a point on the trajectory $\bm{x}_{\tau_\eta} \rightarrow \bm{x}_0^{\eta}$ does not lie on the other trajectory. Additionally, a common point must exist between each denoising trajectory and the SDE trajectory, occurring at $t = \tau_\eta$. Therefore, we can focus on this specific position and only need to prove that the point at $t = \tau_\eta$ on the trajectory $\bm{x}_{\tau_\eta} \rightarrow \bm{x}_0^{\eta}$ does not lie on the trajectory $\bm{x}_{\tau_{\eta'}} \rightarrow \bm{x}_0^{\eta'}$.
\begin{equation}
\bm{x}_t = \sqrt{\alpha_t} \bm{x}_0 + \sqrt{1 - \alpha_t} \bm{\epsilon}, \quad \bm{\epsilon} \sim \mathcal{N}(0, \sigma^2 \bm{I})
\end{equation}

The noise predicted by the teacher model is $\bm{\epsilon}_\phi(x_t)$, where $\bm{\epsilon} - \bm{\epsilon}_\phi(\bm{x}_t) < \bm{\delta}_\phi$. And when the teacher model is perfectly trained, the $\bm{\delta}_\phi = 0$. Denoising from $t$ to $s$ using the DDIM-Solver:

\begin{align}
\hat{\bm{x}}_s^\phi &= \sqrt{\alpha_s} \left(\frac{\bm{x}_t - \sqrt{1 - \alpha_t} \bm{\epsilon}_\theta(\bm{x}_t)}{\sqrt{\alpha_t}}\right) \notag \\
&\quad + \sqrt{1 - \alpha_s - \sigma_t^2} \bm{\epsilon}_\theta(\bm{x}_t) + \sigma_t \bm{\epsilon}_t \\
&= \sqrt{\alpha_s} \left(\frac{\sqrt{\alpha_t} \bm{x}_0 + \sqrt{1 - \alpha_t} \bm{\epsilon} - \sqrt{1 - \alpha_t} \bm{\epsilon}_\theta(\bm{x}_t)}{\sqrt{\alpha_t}}\right) \notag \\
&\quad + \sqrt{1 - \alpha_s - \sigma_t^2} \bm{\epsilon}_\theta(\bm{x}_t) + \sigma_t \bm{\epsilon}_t \\
&= \sqrt{\alpha_s} \left(\bm{x}_0 + \frac{\sqrt{1 - \alpha_t} (\bm{\epsilon} - \bm{\epsilon}_\theta(\bm{x}_t))}{\sqrt{\alpha_t}}\right) \notag \\
&\quad + \sqrt{1 - \alpha_s} \bm{\epsilon}_\theta(\bm{x}_t), \quad \sigma_t = 0
\end{align}
For adding the same noise $\bm{\epsilon}$ from $\bm{x}_0$ to $\bm{x}_s$:
\begin{equation}
\bm{x}_s = \sqrt{\alpha_s} \bm{x}_0 + \sqrt{1 - \alpha_s} \bm{\epsilon}
\end{equation}
The difference between $\hat{\bm{x}}_s^\phi$ and $\bm{x}_s$ is:
\begin{align}
\label{eq:sm7-res-1}
\hat{\bm{x}}_s^\phi - \bm{x}_s &= \frac{\sqrt{\alpha_s} \sqrt{1 - \alpha_t} (\bm{\epsilon} - \bm{\epsilon}_\theta(\bm{x}_t))}{\sqrt{\alpha_t}} 
+ \sqrt{1 - \alpha_s} (\bm{\epsilon} - \bm{\epsilon}_\theta(\bm{x}_t)) \notag \\
&= \frac{1}{\sqrt{\alpha_t}} \left(\sqrt{\alpha_s} \sqrt{1 - \alpha_t} 
- \sqrt{\alpha_t} \sqrt{1 - \alpha_s}\right) (\bm{\epsilon} - \bm{\epsilon}_\theta(\bm{x}_t)) \notag \\
&< C_{t,s} \cdot \bm{\delta}_\phi,
\end{align}
where
\begin{equation}
\label{eq:sm7-res-2}
C_{t,s} = \frac{1}{\sqrt{\alpha_t}} \left(\sqrt{\alpha_s} \sqrt{1 - \alpha_t} - \sqrt{\alpha_t} \sqrt{1 - \alpha_s}\right).
\end{equation}

From Equations \ref{eq:sm7-res-1} and \ref{eq:sm7-res-2}, when the teacher model is perfectly trained, or as the timestep $s$ approaches $t$, the distance between $\bm{x}_s$ and $\hat{\bm{x}}_s^\phi$ decreases. This indicates that the trajectories $( \bm{x}_{\tau_\eta} ) \rightarrow ( \bm{x}_0^{\eta} )$, as well as $( \bm{x}_{\tau_{\eta'}} ) \rightarrow ( \bm{x}_0^{\eta'} )$, are becoming closer to each other as well. However, for an imperfectly trained model, given timesteps $t=\tau_{\eta}$ and $t=\tau_{\eta'}$ where $s\neq t$, the $\bm{x}_{\tau_\eta} \rightarrow \bm{x}_0^{\eta}$ and $\bm{x}_{\tau_\eta'} \rightarrow \bm{x}_0^{\eta'}$ represent two  non-aligned trajectories.
\begin{figure}[htbp]
  \centering
   \includegraphics[width=1.0\linewidth]{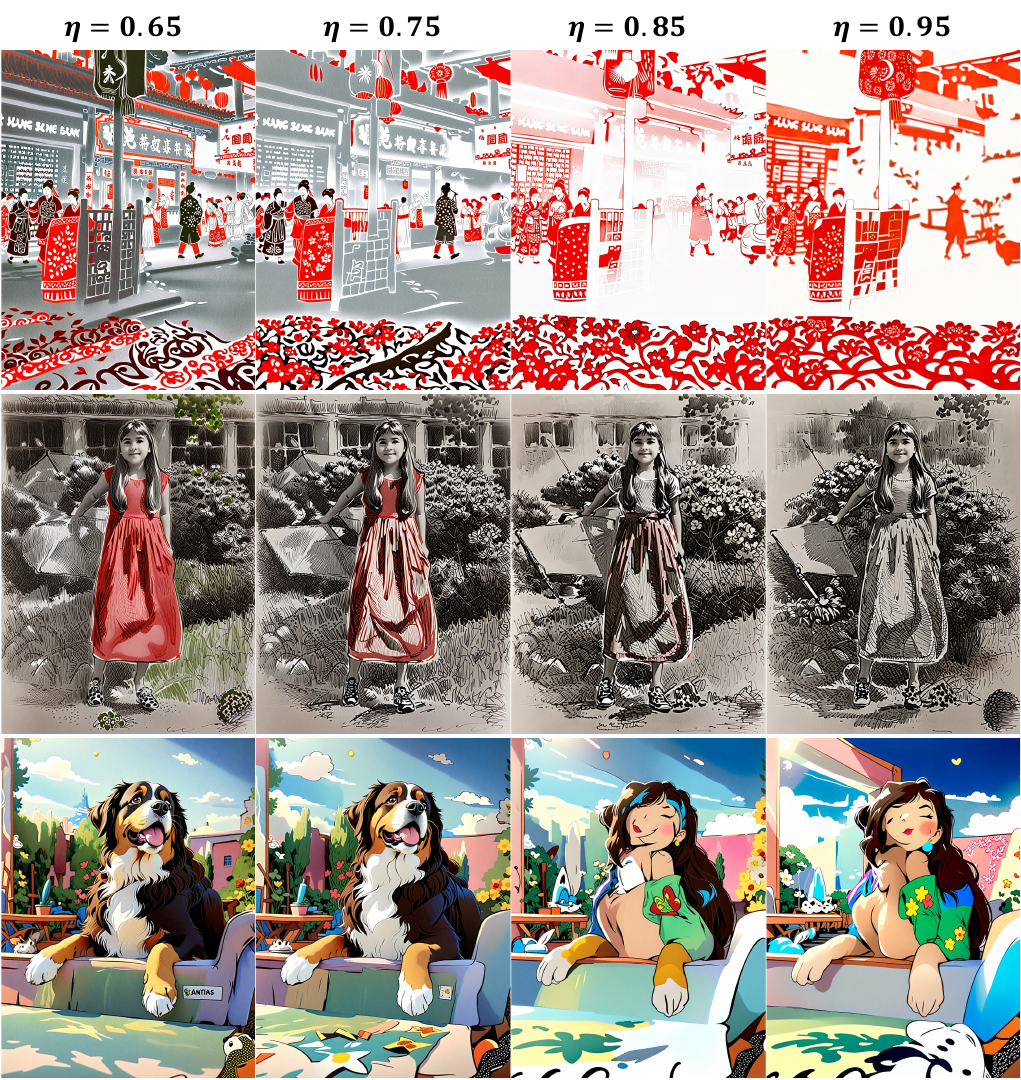}

   \caption{Examples of different strength $\eta$ result with 0.65, 0.75, 0.85 and 0.95.}
   \label{fig:sm-diff-strengths}
\end{figure}

\section{Metrics and Implementation Details}
\label{sec:imple}

\begin{figure*}[htbp]
  \centering
   \includegraphics[width=1.0\linewidth]{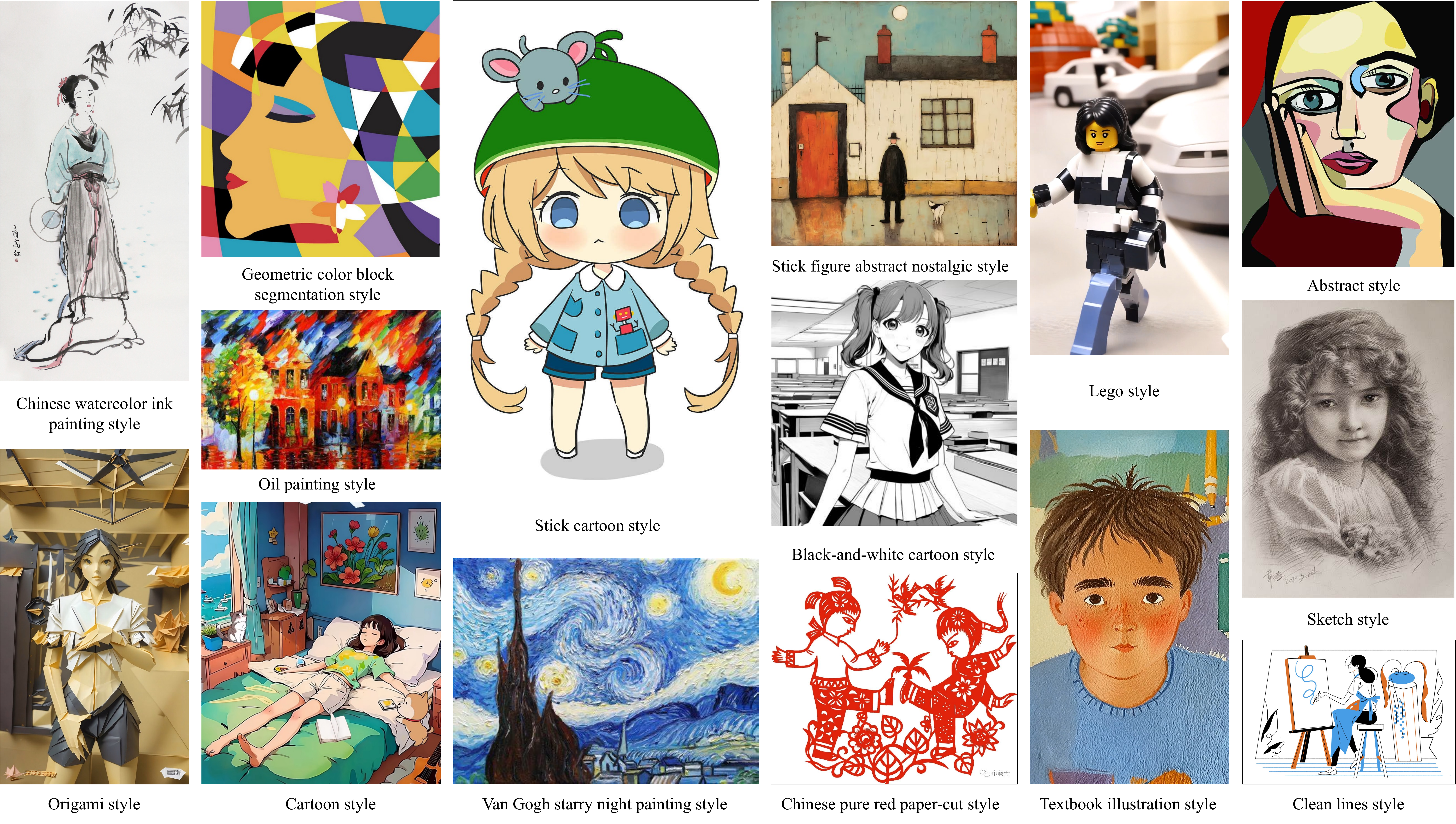}

   \caption{All style images and corresponding prompts.}
   \label{fig:sm8.1}
\end{figure*}
\paragraph{Metrics} In our experiments, we employ the \textit{CSD Score} to evaluate the style similarity between the generated image and the reference style image, the \textit{Aesthetic Score} to assess the overall image quality, and the \textit{Warping Error} to quantify the temporal inconsistency in stylized videos.
The contrastive style descriptors (CSD), denoted as  $l_i = \bm{f}_{ViT}(\bm{x}_i) \in \mathbb{R}^d$ , are utilized to compute similarity scores defined by $s_{i,j} = l_i^T \cdot l_j$. In our experiments, the ViT model employed for this purpose is CSD-ViT-L.
Our aesthetic scores are predicted by the aesthetic predictor v2.5\footnote{\url{https://github.com/discus0434/aesthetic-predictor-v2-5}} which is a SigLIP-based predictor that evaluates the aesthetics of an image in the range of 1 to 10. The aesthetic predictor can evaluate a wider range of image domains, such as illustrations that are more suitable for stylization evaluation.

For each frame $O_t$, we calculate the warping error with frame $O_{t-1}$ as follows:
\begin{align}
E_{pair}(O_t, O_s) = \frac{1}{M_{t, s}}\sum_{i=1}^NM_{t,s}||O_t-W(O_s)||_1  \\
E_{warp}(\{O_t\}_{t=1}^T) = \frac{1}{T-1}\sum_{t=2}^T\{E_{pair}(O_t, O_{t-1})\}
\end{align}
where $M_{t,s}$ is the occlusion map for a pair of images $O_t$ and $O_s$, $N$ is the number of pixels, and $W$ is backward flow with PWC-Net.

\paragraph{Implementation Details} Our image and video test set is collected from the website\footnote{\url{https://www.pexels.com/}} as in previous work~\cite{ku2024anyv2v}. We utilized the TCD scheduler~\cite{zheng2024trajectory} for sampling during inference, and for the comparison methods, we employed their recommended scheduler as indicated on their model pages.

Figure \ref{fig:sm8.1} shows our test-style injection images and corresponding prompt. All experiments use the same style of images and prompts. We inject style information through IP-Adapter, specifically selecting \textit{ip-adapter-plus\_sdxl\_vit-h} model. We only inject image prompts to the 7th block of SDXL UNet according to InstantStyle to keep only style components. 
To make generation results more controllable, we use \textit{Anytest-ControlNet}\footnote{\url{https://huggingface.co/2vXpSwA7/iroiro-lora/tree/main/test_controlnet2}} with depth map and source image. Our input resolution is $1280\times720$, and our negative prompt is related to words about image quality, not style.

For the task of video stylization, we process all videos by extracting frames at a rate of 12 frames per second. The \textit{Hotshot-XL}\footnote{\url{https://huggingface.co/hotshotco/Hotshot-XL}} motion model is employed with a batch size of 8 frames, and an overlap of 4 frames is used to ensure frame consistency.

\section{Additional Generated Samples}
\subsection{Image and Video Comparison Results}
\label{comp}
We show more image and video style transfer results in Figure \ref{fig:sm-image-comp} and Figure \ref{fig:sm-video-comp}. In Figure \ref{fig:sm-image-comp}, we present additional examples where the NFEs are set to 8 and 4 with a CFG of 6. 
Furthermore, we include comparative cases that utilize the recommended CFG settings for different methods. We choose the recommended CFGs according to their model pages and previous work~\cite{wang2024target}. 
Specifically, we use CFG = 1.0 for LCM, TCD, SDXL-Lightning, and MCM, CFG = 1.6 for PCM, CFG = 2.0 for TDD, CFG=6 for Hyper-SD, and CFG=6 for ours. 
Most existing methods use a low CFG scale to mitigate overexposure issues. 
However, for the stylization task, we typically employ a higher CFG scale to preserve the image’s stylistic features better. 
This approach explains why their results often appear less stylized than ours in Figure \ref{fig:sm-image-comp}.

\begin{figure}[htbp]
  \centering
   \includegraphics[width=1.0\linewidth]{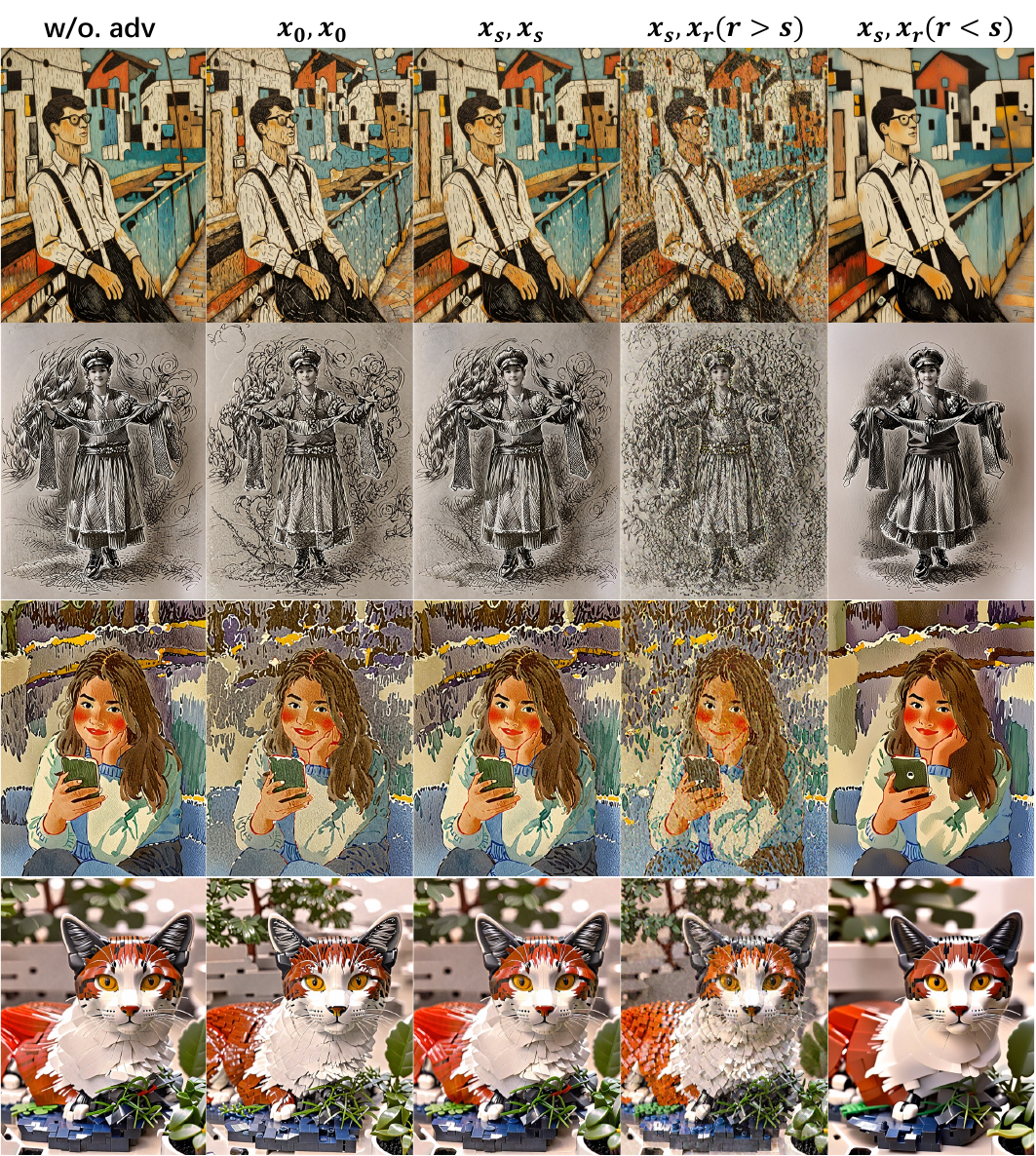}

   \caption{Examples of different adversarial settings.}
   \label{fig:sm-ablation-disc_type}
\end{figure}

\subsection{Ablation Examples}
\label{ablation}

Our paper presents results specifically for the case where the noise strength $\eta$ is set to 0.75. According to the principles underlying the STD method, each value of $\eta$ corresponds to a distinct distillation model. Consequently, we trained four separate models for noise strengths of 0.65, 0.75, 0.85, and 0.95. The results from these models are illustrated in Figure \ref{fig:sm-diff-strengths}. 
Different strengths may be more suitable for different styles. In Figure \ref{fig:sm-diff-strengths}, the first and second rows demonstrate that a higher strength is preferable as it effectively removes the colors from the source images. Conversely, the third row illustrates that a lower strength helps maintain consistency with the source image, preserving more of its original attributes.

Figure \ref{fig:sm-video-comp} presents examples of our video stylization results using equispaced frame extraction. The findings align with image stylization: our method exhibits a more pronounced style, reduced noise, and superior image quality.

In this section, we present ablation studies examples to analyze the impact of different adversarial inputs and the effect of setting  $\gamma$ related to the random target timestep $s$.

\begin{figure}[htbp]
  \centering
   \includegraphics[width=1.0\linewidth]{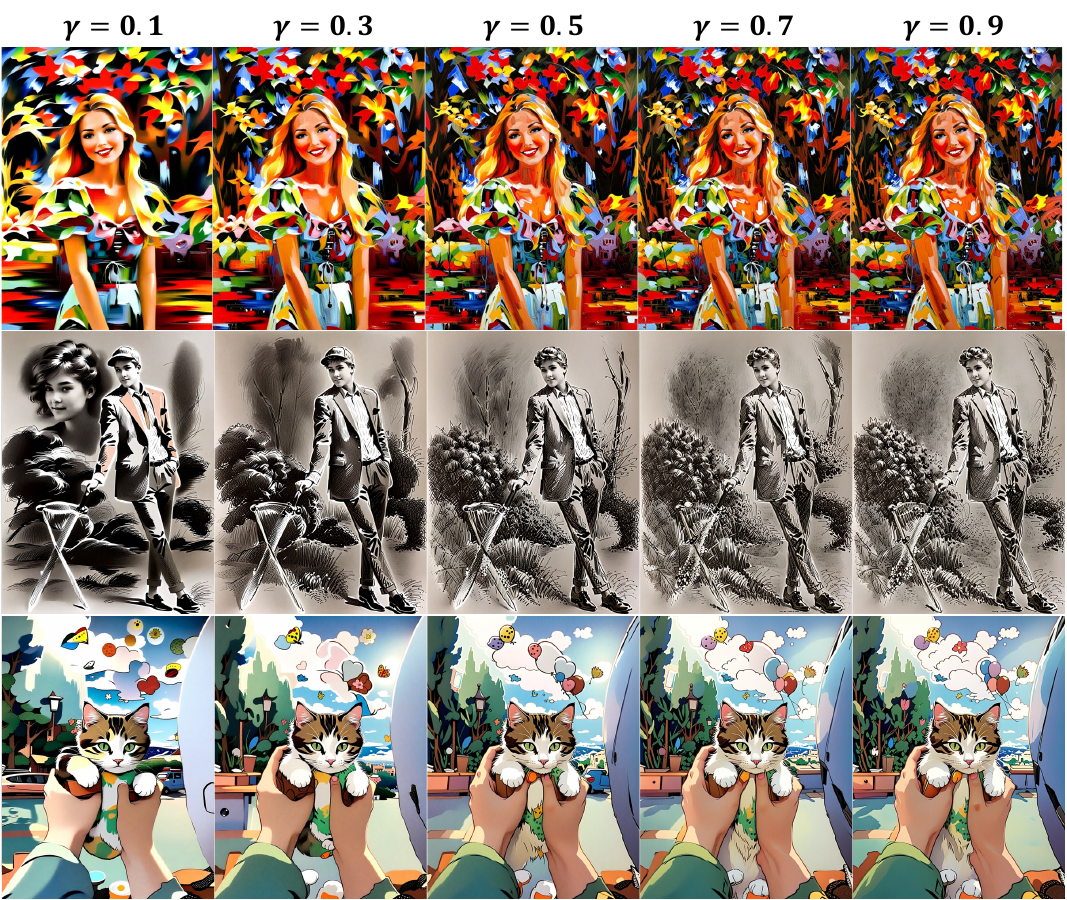}

   \caption{Examples of different $\gamma$ for random target timestep $s$.}
   \label{fig:sm-ablation-strength}
\end{figure}

The adversarial inputs include the model’s predictions and the real images. For example, we denote these as $\bm{x}_s, \bm{x}_r$, where $\bm{x}_s$ represents the prediction at timestep $s$, and $\bm{x}_r$ refers to the real images with noise added to timestep $r$.

Figure \ref{fig:sm-ablation-disc_type} shows different adversarial input settings results.
First, compared to the case without adversarial loss, the settings $\bm{x}_0, \bm{x}_0$ and $\bm{x}_s, \bm{x}_s$ produce images with more detailed features. 
However, increased detail often comes at the cost of higher noise levels. 
Using the $\bm{x}_s, \bm{x}_r$ setting where $r > s$ results in images that are overly noisy, leading to lower image quality. 
On the other hand, the $\bm{x}_s, \bm{x}_r$ configuration where $r < s$ yields images with reduced noise, as well as higher saturation and contrast, resulting in cleaner and higher-quality outputs.

We also discuss the influence of different target timestep range rates $\gamma$ on predicted images. The range rate $\gamma$ determines the lower bound of the value range of target timestep $s$. As shown in equation \ref{eq:4.7}, when the $\gamma$ is too small, too little detail is presented, and there is a phenomenon of overexposure, which ultimately distorts the style. In practice, we recommend using a larger $\gamma$, such as 0.7 or 0.9, for better style retention.

\begin{figure*}[htbp]
  \centering
   \includegraphics[width=1.0\linewidth]{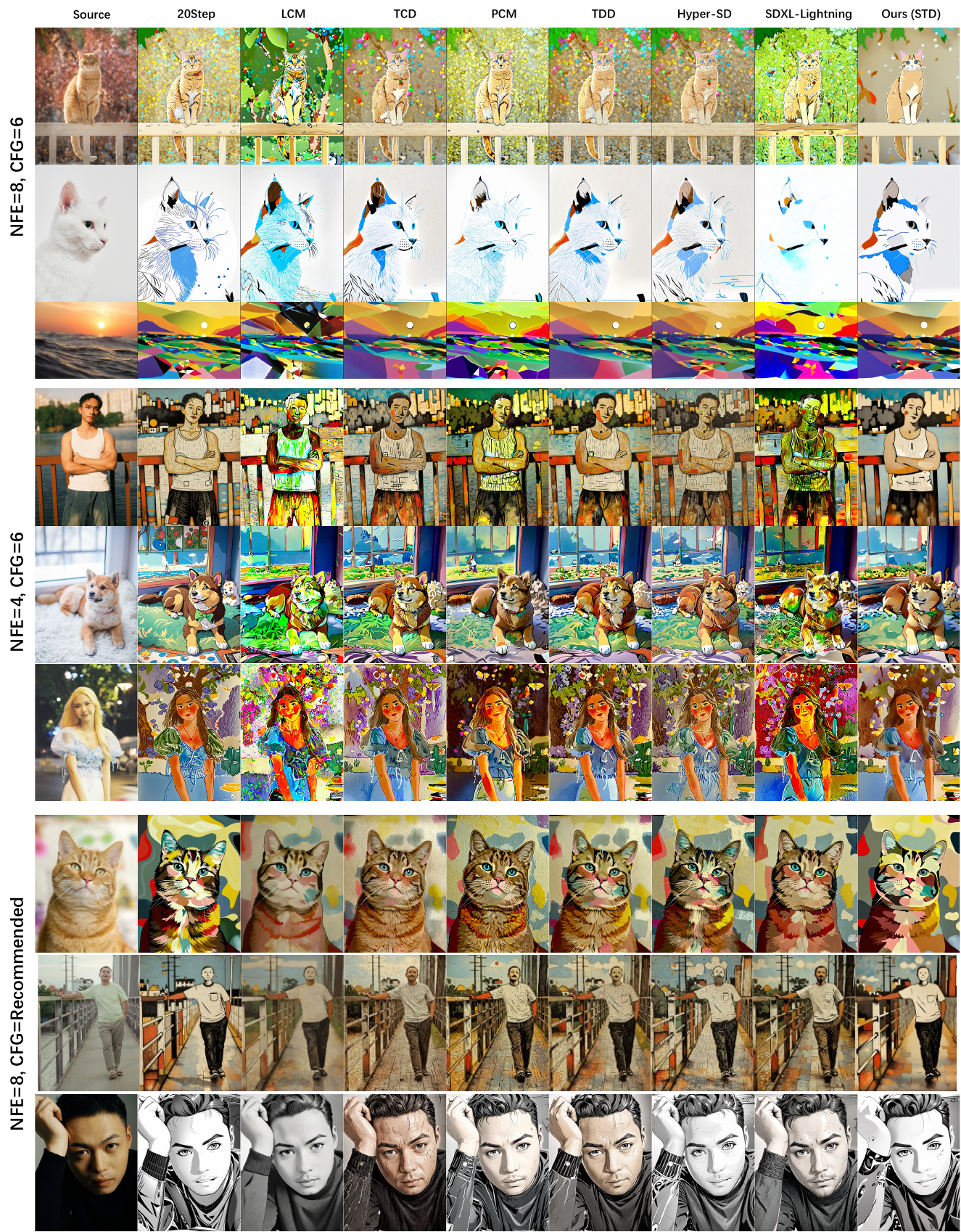}

   \caption{More comparison results in different NFEs.}
   \label{fig:sm-image-comp}
\end{figure*}


\begin{figure*}[htbp]
  \centering
   \includegraphics[width=1.0\linewidth]{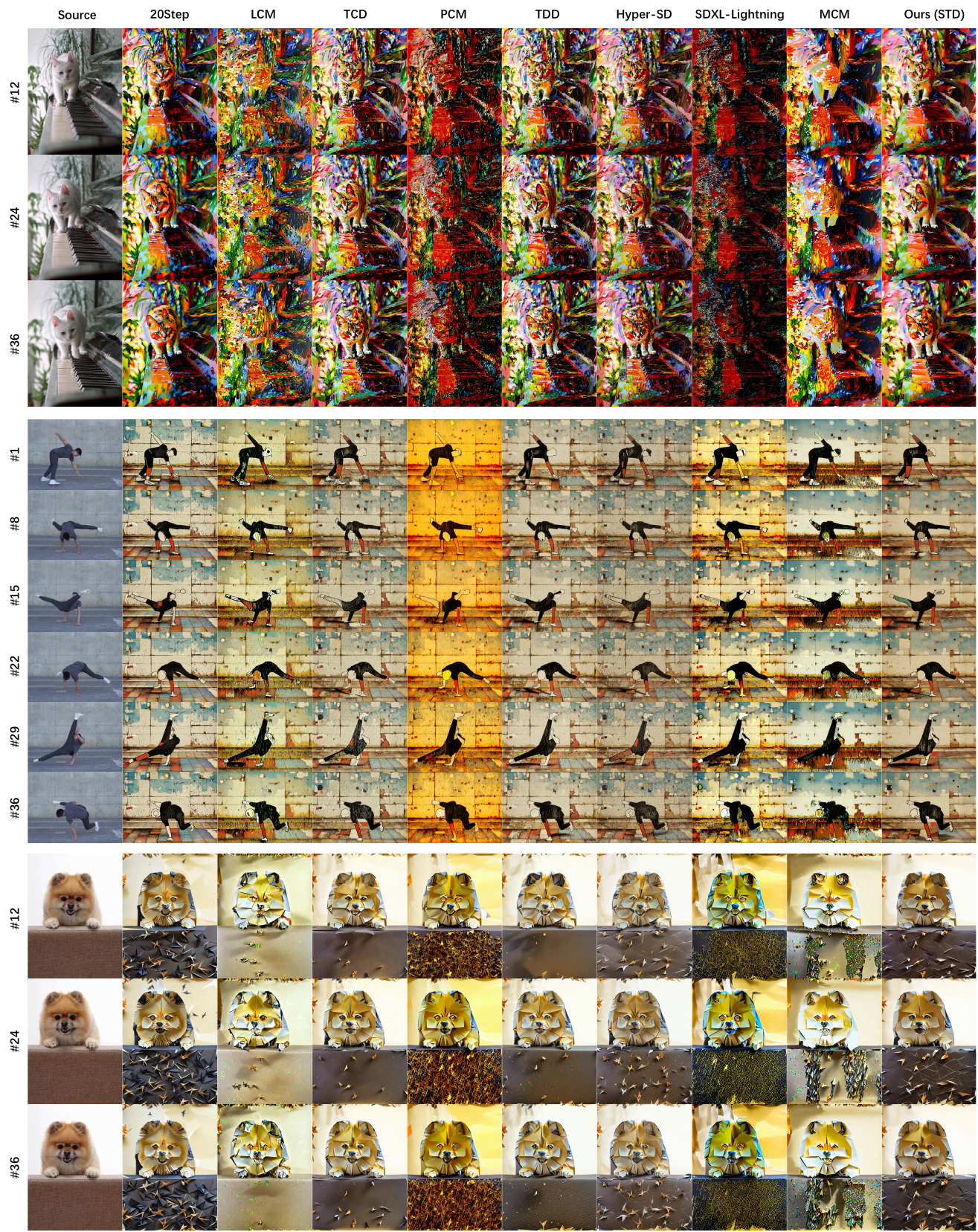}

   \caption{Video style transfer results in NFEs=8 and CFG=8.}
   \label{fig:sm-video-comp}
\end{figure*}


\end{document}